\documentclass{article}
\usepackage{amsmath,amssymb}
\usepackage{algorithmicx}
\usepackage{graphicx}
\usepackage{algorithm}
\usepackage{array}
\usepackage{algpseudocode}
\usepackage{subcaption}
\usepackage{tikz}
\usepackage{cite}
\usepackage{authblk}
\usepackage[colorinlistoftodos]{todonotes}
\usetikzlibrary{shapes.geometric}
\usetikzlibrary{positioning}
\tikzstyle{var} = [rectangle, minimum width=3cm, minimum height=1cm, text centered, draw=black]
\tikzstyle{process} = [circle, text centered, draw=gray]
\tikzstyle{changedynamic} = [circle, text centered, draw=black]
\tikzstyle{changepoint} = [circle, text centered, draw=blue]
\tikzstyle{input} = [rectangle, minimum width=3cm, minimum height=1cm, text centered, draw=black]
\tikzstyle{arrow} = [thick,->,>=stealth]

\usepackage{cite}
\usepackage{graphicx}
 
\graphicspath{ {../img/} }
\usepackage{color}

\begin{document}
		\title{A Change-dynamic Model for the Online Detection of Gradual Change}
		\author{Chris Browne, cnb49@cornell.edu}
        \affil{\vspace{-2ex}Center for Applied Math, Cornell University}
        \date{\vspace{-5ex}}
        \maketitle

\begin{abstract}

\noindent Changes in the statistical properties of a stochastic process are typically assumed to occur via change-points, which demark instantaneous moments of complete and total change in process behavior.  In cases where these transitions  occur gradually, this assumption can result in a reduced ability to properly identify and respond to process change.  With this observation in mind, we introduce a novel change-dynamic model for the online detection of gradual change in a Bayesian framework, in which change-points are used within a hierarchical model to indicate moments of gradual change onset or termination.  We apply this model to synthetic data and EEG readings drawn during epileptic seizure, where we find our change-dynamic model can enable faster and more accurate identification of gradual change than traditional change-point models allow.

\end{abstract}

\section{Introduction}

Natural processes may undergo transient periods of nonstationarity which produce lasting change in process behavior across time.  When driven by exogeneous influences these changes can be challenging to predict in advance.  To circumvent this challenge, works in online (sequential) change detection aim to deduce the occurrence of change in process behavior as it occurs via direct observation of an online data stream.  While such changes in process behavior are most commonly modeled via change-points, in which the parameters and/or densities defining an associated process model are assumed to undergo an abrupt and instantaneous transition, changes in the behavior of some processes may occur gradually, taking time to reach their full effect. In such cases change-point models may be ill suited, producing either inaccurate estimates for the timing of these changes or, when this gradual change occurs slowly and when change detection is performed concurrently with model estimation, failing to properly detect change occurrence, as we show empirically in Section \ref{sec:exp}. This effect can have a significant impact in application, where automated system controls may not be appropriately applied at the correct times, and can result in inaccurate models of process behavior during and after this gradual change.

\

\noindent  With this observation in mind we introduce a novel ``change-dynamic" model for the online detection of gradual change, in which change-points are used within a hierarchal model to demark the onset and/or termination of successive gradual changes in process behavior.  To the best of our knowledge, this is the first work to explicitly consider the modeling and detection of gradual change in an online setting.  We find that when applicable the performance of this model is superior to comparable change-point methods, producing more precise estimates of process behavior, enabling both faster and more accurate detection of process change.  Moreover, by treating the detection of gradual change in a Bayesian framework, we also allow for the online inference of how a particular system undergoes gradual change, even when prior knowledge of these change-dynamics are uncertain.

\

\noindent In Section \ref{sec:relatedworks}, we introduce related works and discuss how they relate to our change-dynamic model. In Section \ref{sec:process} we introduce some preliminary notation for our data and process model and show in section \ref{sec:model} how gradual changes in the behavior of this process model are addressed via our change-dynamic model.  Inference in this model is performed via a particle filter, whose implementation is discussion in Section \ref{sec:inference}.  We then show how prediction and change detection can be performed in the context of this inference scheme in Sections \ref{sec:prediction}, \ref{sec:changedetection} and discuss the computational cost associated with these operations in Section \ref{sec:computability}.  Finally, in Section \ref{sec:exp} we apply our method to synthetic and real data and assess our results.

\section{Related Works}
\label{sec:relatedworks}

\noindent  The study of online (sequential) change detection begins with the works of Page \cite{page54,page57} and Shiryaev \cite{shiryaev61b,shiryaev61a}, who considered the online detection of an abrupt shift in the mean of a Wiener process.  While Page treated the location of this change-point as a non random quantity to be estimated via a sequence of likelihood ratios, Shiryaev adopted a Bayesian approach in which the location of this change-point was treated via a geometric prior, with change declared via a sequence of posterior thresholds.  Their procedures, the frequentist CUSUM and Bayesian Shiryaev protocols, alongside modest extensions and modifications of these procedures, have since been shown to be to be optimal for many online settings (see e.g. Baron \cite{baron2006asymptotic}, Tartakovsky \cite{tartakovsky2005general,tartakovsky2008asymptotically}, Veeravalli \cite{veeravalli2014quickest}), meaning that they minimize the average delay of change-point detection (ADD) subject to an upper bound on either the probability of false alarm (PFA), or, for frequentist approaches, the reciprocal mean time to false alarm conditional on change-point non occurrence (FAR).  These protocols thus remain at the heart of most modern methods for online change detection, including our own, in which we detect gradual change via a particle filtered approximation a to classical Shiryaev protocol.

\



\

\noindent  In contrast to the assumptions made in these pioneering works, knowledge of the parameters defining pre and/or post-change process models are often unknown in advance, necessitating online model inference or estimation concurrent with change detection.  While Adams, Mackay \cite{adams2007bayesian} consider the case in which inference over both model parameters and change-point locations can be performed analytically online, sliding windows may be used be used in conjunction with generalized likelihood ratio tests (GLRT) over both pre and post change model parameters as in Lai, Xing \cite{lai10}.  When either a pre or post-change model is specified one might conduct change detection under a worst case scenario corresponding to the infimum of the resulting GLRT, as in \cite{mei06}. While these methods are effective when estimating model parameters in tandem with change-point detection, they may be less suited to to the online detection of gradual change.  The use of GLRTs or Bayesian update schemes which do not explicitly incorporate prior knowledge of gradual change occurrence may result in the gradual loss of the appropriate pre-change model as it is tuned to better reflect data drawn from the beginning of a gradual transition into post change process behavior.  This can result in challenged comparison to any post-change model and in turn heightened detection delay or failure to detect change. In contrast, by explicitly incorporating prior knowledge of gradual change occurrence, we find our change-dynamic model is less sensitive to this effect.   Although equally applicable to the identification of gradual change, assumption of a ``worst case" scenario can increase detection delay and fails to provide estimates of model parameters which might be necessary for diagnostic, control, or response purposes.

\

\noindent The change-points of traditional change-point models can also be interpreted as demarking moments at which a hidden state variable modulating the parameters of a process model abruptly change.  Change-point models can thus be viewed from the context of regime switching models and bear similarities to classical works in this domain such as the threshold autoregressive model of Tong \cite{tong1978threshold} and the Markov switching autoregressive model of Hamilton \cite{hamilton1989new}, modulo any emphasis on explicit change-detection.  In fact, our change-dynamic model soon to be introduced can be interpreted as an instance of Hamilton's model, but whose outputs are the parameters of a model for some potentially nonstationary process and whose states signify the type of gradual transition, or lack thereof, in effect at each time.  This state based formulation allows for the classification of how a given system is or is not gradually changing over time.  Such classification can be of interest in applications where we may wish to apply distinct controls or raise distinct alarms depending on what type of gradual change has been detected.  Such frameworks have previously been considered in the case of online change-point detection, e.g. by Draglia et al. \cite{draglia1999multihypothesis}
and Tartakovsky \cite{tartakovsky1998asymptotic}, with the later of these works informing our own change-detection procedure.

\

\noindent Though online detection of gradual change has, to the best of our knowledge, been ignored by prior works, the identification of a pre-specified number of gradual changes in retrospective settings has received moderate attention. Terasvirta \cite{terasvirta1994specification} and Luukkonen et al. \cite{luukkonen1988testing} introduced the Smooth Transition Autoregressive (STAR) model, in which retrospective inference of a parametrically defined gradual transition between two models was addressed.  Similarly, Pastor-Barriuso et al. \cite{pastor2003transition} and Hout et al. \cite{van2011smooth} applied the work of Bacon and Watts \cite{bacon1971estimating} to retrospectively infer a parametric, gradual transition between model regimes.  More recent work by Wilson \cite{wilson13} introduced the change-surface kernel as a means of retrospectively modeling a pre-specified number of gradual transitions in the covariance structure of a Gaussian process.  Herlands et al. \cite{herlands16} considered an application of this kernel to facilitate the retrospective identification of a pre-specified number of change-surfaces (e.g. gradual model transitions), while Lloyd et al. \cite{lloyd2014automatic} applied this kernel in the context of a broader model discovery and annotation procedure which considered the retrospective, automated, identification and labeling of an unknown number of gradual changes in a Gaussian process model.

\

\

\noindent While these related works form a strong foundation for the modeling of nonstationary processes, they either (i) assume that changes in process behavior occur instantaneously via change-points, (ii) assume a pre specified number of gradual changes in process behavior, (iii) are retrospective, and/or (iv) eschew any explicit modeling or detection of process change.  In this work, we will aim to detect and classify an unknown number of gradual changes in process behavior as they occur online, while inferring the parameters defining these gradual changes and post-change process behavior accordingly.

\section{Methodology}
\label{sec:methods}

\subsection{Preliminary Notation and Process Model}
\label{sec:process}

Our work begins with the consideration of a stochastic process $X(t)$, with $X(t) \in \mathcal{R}$ for $t=1,...,T$.  We assume that observations of this process $x_1,...,x_T$ are viewed sequentially (online), with the index $t$, henceforth referred to as time, used to indicate the ordering of this observation.  Throughout this text, we use the shorthand  $X = X(\cdot)$ used as shorthand notation to refer to this process at some arbitrary time.  At various, unknown, times, we assume the process  $X$ will undergo gradual changes in behavior which we aim to describe via a hierarchal model consisting of two key components: The first, a process model for $X$, and second, a change-dynamic model by which the parameters, and thus dynamics, of this process model may gradually vary over time.

\

\noindent  In this work, we consider simple autoregressive process models of form:

\begin{subequations}
\begin{align}
X_{t+1} = \mathbf{\alpha}_t^T\tilde{\mathbf{x}}_t + \mu_t + \sigma_t \epsilon_t \\
\epsilon_t \sim \mathcal{N}(0,1),
\label{eq:obsfun}
\end{align}
\end{subequations} 

\noindent where $\tilde{\mathbf{x}}_t = \mathbf{x}_{t-p+1:t}$ denotes the past $p$ lagged observations of $X$ at time $t$ and with $\tilde{\mathbf{x}}_0$ assumed known or sampleable from an initial density $p_{\mathbf{\tilde{x}}_0}$.  While the parametric form of Equation \ref{eq:obsfun} assumes that the dynamics of $X$ will be well represented by a linear autoregressive model, the exact behavior of Equation \ref{eq:obsfun} at any given time will be controlled by the values adopted by a collection of time-varying parameters. For conciseness, we concatenate these parameters into a single vector $\boldsymbol\theta_t = [\mathbf{\alpha}_t;\mu_t;\log(\sigma_t)] \in \mathcal{R}^m$.  Gradual changes in the behavior of $X$ will then be modeled via gradual changes in the values of $\boldsymbol\theta = \boldsymbol\theta(\cdot)$, whose time evolution is treated via our change-dynamic model.

\subsection{Change-dynamic Model}
\label{sec:model}

\noindent  Our change-dynamic model itself consists of two main components.  The first is a collection of independent, autoregressive, priors over the components of $\boldsymbol\theta$, which we refer to collectively as a change-dynamic:

\begin{subequations}	
\label{eq:transfun}
\begin{align}
\theta_{t+1}^i =  \tilde{\boldsymbol\theta}_{t}^i + \nu_{t+1}^i  + \gamma_{t+1}^i \omega_{t}^i,  \\
\omega_{t}^i \sim \mathcal{N}(0,1).
\end{align}
\end{subequations}

\noindent Here $\theta_{t}^i$ denotes the value of the $i$th element of $\mathbf{\theta}_t$, for $i=1,...m$,
$\tilde{\mathbf{\theta}}_{t}^i = \theta_{t-q+1:t}^i$ denotes the past $q$ lagged values of $\theta^i$ at time $t$, and each $\tilde{\mathbf{\theta}}_{0}^i$ is assumed known or sampleable from an initial density $p_{\tilde{\mathbf{\theta}}_{0}^i}$.  The parameters of this change dynamic can be understood as follows: each $\nu_{t+1}^i$ denotes the mean rate of gradual change in $\theta^i$ at time $t$, while each $\gamma_{t+1}^i$ allows for small deviations about this linear dynamic.  For conciseness, we again concatenate the parameters of this change-dynamic Equation \ref{eq:transfun} into a single vector $\boldsymbol\phi_{t+1} =  [\nu_{t+1}^{1},...\nu_{t+1}^m, \gamma_{t+1}^{1},...,\gamma_{t+1}^m]^T$.  Throughout this text, we refer to these quantities as our change-dynamic parameters, for which we again employ a shorthand $\boldsymbol\phi = \boldsymbol\phi(\cdot)$ to refer to these quantities at any arbitrary time.

\

\noindent Beneath this change-dynamic, we then consider a hidden Markov process whose switching times will coincide with the successive onset and/or termination of gradual changes in model behavior.  The primary action of this process will then be to modulate the parameters $\boldsymbol\phi$ of Equation \eqref{eq:transfun} which define the dynamic by which the process model parameters $\boldsymbol\theta$ gradually change over time.

\begin{subequations}
\label{eq:hmm}
\begin{align}
\boldsymbol\phi_{t+1} \sim 
\begin{cases}
Uniform(\boldsymbol\eta_j^{min},\boldsymbol\eta_j^{max}), \text{ if } R_{t+1} = 0, S_{t+1} = j, \text{ for } j=0,...,K \\
\delta(\cdot;\boldsymbol\phi_{t}) \text{ else } 
\end{cases} \\
S_{t+1} \sim 
\begin{cases}
Categorical(P_S(S_{t})), \text{ if } R_{t+1} = 0, S_t = j, \text{ for } j=0,...,K \\
\delta(\cdot;S_{t}) \text{ else }
\end{cases}
\end{align}
\end{subequations}

\begin{equation}
\label{eq:hazard}
R_{t+1} = (R_t + 1)(1-I(R_t;R_t \geq t-\tau)Binom(p(S_t))),
\end{equation}

\noindent where $\delta(\cdot;z)$ denotes a degenerative distribution centered at $z$.  We assume that $S_0 = s_0$  is known, that $R_0=0$, and that either $\boldsymbol\phi_0 \sim Uniform(\boldsymbol\eta_{s_0}^{min},\boldsymbol\eta_{s_0}^{max})$ or known.  Here, $\tau$ is an indicator of the most recent time at which gradual change onset and/or termination was identified in our model, with initial $\tau=0$.  During inference, $\tau$ will be updated as changes are declared via a scheme discussed in Section \ref{sec:changedetection}.

\

\noindent To more fully describe how this change-dynamic model is used to describe gradual changes in process behavior, we begin at the highest level of our model hierarchy \eqref{eq:hazard} wherein a runlength variable $R_t \geq 0$ measures the time since onset of the current change-dynamic (i.e. gradual change) active at time $t$ or the time since onset of process stationarity.  For the remainder of this text, we refer to such moments as change-points, noting that the primary purpose of $R = R(\cdot)$ will then be to act as a signifier of change-point occurrence and which is used to compute our change-point statistic in Section \ref{sec:changedetection}.  The time evolution of $R$ given in Equation \eqref{eq:hazard} can be understood as specifying a regenerative process. Modulo the interference of an indicator variable, the time between any two change-points is presumed to be geometrically distributed, with incremental rate $p(S)$ dictated by the current state of our system.  Once a change-point has occurred, this indicator then prevents the occurrence of additional change-points until said change has been detected, corresponding to an assumption that gradual changes should be detected one at at time, which is further discussed and motivated in Section \ref{sec:changedetection}.  At such times $t$ of change declaration, we update $\tau = t$, thus regenerating the process \ref{eq:hazard} and allowing for the subsequent occurrence and identification of future change-points.  

\

\noindent Below this runlength we consider a Markovian state $S_t \in {0,1,..,K}$ which specifies the type of gradual change in model behavior currently in effect at time $t$, with $S = S(\cdot) = 0$ reserved to indicate periods of model stationarity. We again motivate this categorical formulation of $S$ by the need to, in many applications, identify and respond to several distinct behavioral changes which a process might undergo and to facilitate the re-identification of model stationarity following the termination of a gradual change.  We now motivate our state dependent transition rates $p(S)$ as allowing for different types of gradual transitions to exhibit different mean runlengths.  Analogously to $R$,  $S_{t+1} = S_t$ remains fixed between change-points.  At change-points, a new-post change state is proposed conditional on the pre-change state $S_t$ via an embedded state transition matrix $P_S$, with $P_S(S_t)$ denoting the vector of transition probabilities away from state $S_t$. In specification of $P_S$, we assume the total number of possible types of gradual change $K$ is specified in advance via user prior knowledge.  Though we consider in this paper exclusively uniform transition matrices, this assumption could be easily modified based on domain expertise or training data informing what sequences of gradual changes are likely to occur in practice.

\

\noindent  Finally, at the lowest level of our change-dynamic model, a collection of autoregressive priors over the parameters of our process model $\mathbf{\theta}$, or change-dynamic, are used to enact gradual changes in process model behavior.  The precise manner in which $\mathbf{\theta}$ evolves at any given time $t$ under Equation \ref{eq:transfun} is controlled by the values of the change-dynamic parameters $\mathbf{\phi}_t$, with each $\nu_t^i$ controlling the mean rate of gradual change in $\theta^i$ at time $t$ and with $\gamma_t^i$ allowing for deviations from this linear dynamic. $\mathbf{\phi}$ will remain fixed between change-points, as enforced by  during which time $\mathbf{\theta}$ will gradually change under a fixed dynamic or remain stationary.  At change-points, new values for $\mathbf{\phi}$ are proposed via sampling from a uniform prior associated with this type of gradual change, thereby enacting a new dynamic for the gradual evolution of $\mathbf{\theta}$.

\

\noindent These uniform priors, Equation \eqref{eq:hmm}, alongside their associated bounds $\mathbf{\eta}_i^{min},\mathbf{\eta}_i^{min}$, are used to afford uncertainty in prior specification of each change-dynamic, whose exact rate of change may be challenging to precisely specify in advance.  In the ideal case, selection of these bounds could be performed through inspection of ground truth transitions in model behavior; For each type of gradual change considered for a given application, one could sequentially fit a process model to data sampled during this transition, producing a sequence of parameter values on which a change-dynamic could be directly fit.  With repeated realizations of such data, uniform bounds for the parameters $\nu^{1:m},\gamma^{1:m}$ could then be generated, with this process repeated for each type of gradual change which may occur in application.  However, in the case where such information is unavailable we concede that this specification must be ultimately informed by prior knowledge, and later highlight how such specification might be performed in Section \ref{sec:exp}.  We briefly note our usage of uniform priors bears some similarity to the range of plausible values for parameters defining an unknown pre or post-change process model discussed in \cite{lai10,mei06}, though we will aim here to infer the appropriate values of each $\nu^i_t,\gamma^i_t$ rather than taking a single point estimate for these quantities.

\

\noindent Noting Equation \ref{eq:transfun}, we assume in this work that changes in model behavior will occur via a linear dynamic, e.g., that the parameters of our process model may gradually vary via a linear dynamic or remain fixed, while adopting a collection of uniform priors over the parameters $\mathbf{\phi}$ of this change-dynamic in Equation \ref{eq:hazard}.  We emphasize that even linear change-dynamics allow for considerably more sophisticated treatment of gradual change than traditional change-point models, which can be recovered in our model by adopting a piecewise constant change-dynamic: $\mathbf{\theta}_{t+1} = \mathbf{\nu}_{t+1}$.  However, in cases where one might wish to adopt a more complex model of gradual change, or employ alternative priors over the parameters defining these gradual changes, one could still apply our inference scheme discussed in Section \ref{sec:inference}, and supports straightforward extensions of our model which could serve as a direction of future works.   

\

\begin{figure}[H]
\begin{tikzpicture}[node distance=2cm]

\node (rm) [draw,circle] {$R_{0:t-1}$};
\node (r) [draw,circle,above of=rm] {$R_{t}$};
\node (sm) [draw,circle,right of=rm] {$S_{0:t-1}$};
\node (s) [draw,circle,above of=sm] {$S_{t}$};
\node (phim) [draw,circle,right of=sm] {$\mathbf{\phi}_{0:t-1}$};
\node (phi) [draw,circle,above of=phim] {$\mathbf{\phi}_{t}$};
\node (thetam) [draw,circle,right of=phim] {$\mathbf{\theta}_{0:t-1}$};
\node (theta) [draw,circle,above of=thetam] {$\mathbf{\theta}_{t}$};
\node (xm) [draw,rectangle, right of=thetam] {$\mathbf{x}_{0:t-1}$};
\node (x) [draw,rectangle,above of=xm] {$\mathbf{x}_{t}$};
\node (xp) [draw,rectangle,above of=x] {$\mathbf{x}_{t+1}$};

\draw [arrow] (rm) -- (sm);
\draw [arrow] (sm) -- (rm);
\draw [arrow] (sm) -- (phim);
\draw [arrow] (phim) -- (thetam);
\draw [arrow] (thetam) -- (x);
\draw [arrow] (theta) -- (xp);

\draw [arrow] (sm) -- (r);

\draw [arrow] (r) -- (s);
\draw [arrow] (s) -- (phi);
\draw [arrow] (phi) -- (theta);

\draw [arrow] (rm) -- (r);

\draw [arrow] (sm) -- (s);

\draw [arrow] (phim) -- (phi);

\draw [arrow] (thetam) -- (theta);

\draw [arrow] (xm) -- (x);
\draw [arrow] (x) -- (xp);

\end{tikzpicture}
\caption{Summary Plate Diagram for Change-dynamic Model.  Model variables treated as random are displayed as circles while observations are marked as rectangles.}
\label{fig:plate}
\end{figure}
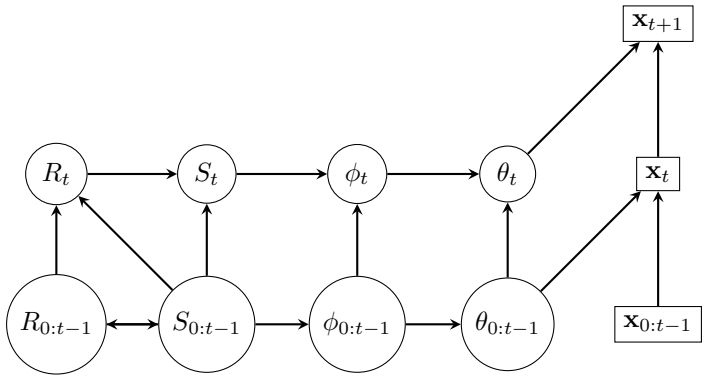

\noindent  Having introduced our full change-dynamic model, we offer here a brief summary of our model, and display a summary plate diagram in Figure \ref{fig:plate}.  At the lowest level of our hierarchy, a process model with time-varying parameters $\boldsymbol\theta$ is used to describe the dynamics of a nonstationary stochastic process $\mathbf{X}$.  Gradual changes in process behavior are then modeled via gradual changes in the values of $\boldsymbol\theta$, whose time evolution is treated via a collection of autoregressive priors, or change-dynamic, \ref{eq:transfun}.  The parameters $\boldsymbol\phi$ of this change-dynamic are dependent on a hidden state $S$ which indicates the type of gradual change in process behavior, or lack thereof, in effect at any given time.  Change-dynamic parameters $\boldsymbol\phi$ and state $S$ are fixed for the duration of any gradual change or period of model stationarity, during which time process model parameters $\boldsymbol\theta$ evolve gradually under a fixed autoregressive dynamic while a runlength variable $R$ increases monotonically.  The onset of a new evolutionary dynamic for $\boldsymbol\theta$ at time $t$ will then coincide with the event $R_t=0$ (change-point occurrence) and the proposal of a new post-change state $S_t$.  At such times, we set $\tau = t$, allowing for the identification of later change-points.  We then propose new values for the change-dynamic parameters $\boldsymbol\phi$ via sampling from a uniform density encoded by this post-change state, thereby enacting a new dynamic by which $\boldsymbol\theta$ will gradually change over time under the change-dynamic \ref{eq:transfun}.

\

\subsection{Posterior Inference}
\label{sec:inference}

While our model posterior is in general analytically intractable, approximate inference can be easily accomplished through the use of a sequential importance resampling scheme (i.e. particle filter, bootstrap filter) \cite{doucet2001introduction}.  Concatenating all model parameters available at time $t$, $\mathbf{p}_t= (\boldsymbol\theta_{-q+1:t-1},\boldsymbol\phi_{0:t-1},S_{0:t-1},R_{0:t-1})$, we assume our model posterior at time $t$ can be written as a weighted sum of $N$ delta functions centered at particles $\mathbf{p}_t^i, i=1:N$:

\begin{equation}
\label{eq:particlepost}
P(\mathbf{p}_t|X_{-p+1:t}) \simeq \sum_{i=1}^{N} \frac{1}{N} \delta(\mathbf{p}_t;\mathbf{p}_t^i), 
\end{equation}

\noindent where $\mathbf{p}_t^i = (\boldsymbol\theta_{-q+1:t-1}^i,\boldsymbol\phi_{0:t-1}^i,s_{0:t-1}^i,r_{0:t-1}^i)$ contains the parameter values associated with our $i$-th particle.  Inference begins by generating an initial posterior of form \ref{eq:particlepost} via sampling of $N$ particles $\mathbf{p}_0^i$  from the appropriate initial densities, which may be degenerative, alongside an initial $\tilde{x}_0$ which is assumed given.  Given such an approximation at time $t$,  posterior updates at time $t+1$ are performed by first naively proposing updates to particle parameters $r_{t+1}^i, s_{t+1}^i,\boldsymbol\phi_{t+1}^i,\boldsymbol\theta_{t+1}^i$ via sampling from the appropriate priors specified by Equations \ref{eq:transfun},\ref{eq:hmm},\eqref{eq:hazard}. Following extension, 
particles are then assigned a score based on their one step ahead likelihood for the newly observed datum $x_{t+1}$: $w_{t+1}^i = P(X_{t+1}|x_{-p+1:t},p_{1:t+1}^i)$.  Particle scores are then normalized to produce weights encoding the relative strength of each particle: $W_{t+1}^i = \frac{w_{t+1}^i}{\sum_{i=1}^{N} w_{t+1}^i}$.  Following normalization, posterior updating then concludes by resampling $N$ particles, with replacement, from the current particle set, with sampling weights given by the normalized weights $W_{t+1}^i$.  Beyond its ease of use, we further motivate our choice of inference scheme as enabling the extension of our change-dynamic model to alternative observation likelihoods, change-dynamics, and change-dynamic parameter priors which might aid in future extensions of our relatively simple model to future applications.

\subsection{Change Detection}
\label{sec:changedetection}

\noindent Given our filtered posterior at time $t$, change detection can then be easily performed by direct evaluation of posterior support for or against the occurrence of a new change-point.  Due to the specification of our runlength prior \ref{eq:hazard} and inference scheme, these quantities will correspond to the number of particles whose runlengths have collapsed to 0 at exactly one point since the prior moment at which change was declared and the number of particles in which no new change-point has occurred.  In particular, we compute:

\begin{equation}
\label{eq:changedetection}
    Z_{t} = \frac{P(R_t < t-\tau|x_{-p+1:t})}{P(R_t \geq t-\tau|x_{-p+1:t})}
    \simeq \frac{\sum_{i=1}^N I(r_{t}^i < t-\tau)}{\sum_{i=1}^N I(r_{t}^i \geq t-\tau)},
\end{equation}

\noindent where $I(A)$ is here used to refer to a indicator function on the event $A$.  Equation \ref{eq:changedetection} can be understood as a particle filtered approximation to a traditional Shiryaev protocol. We then declare change at time $t$ whenever $Z_t > h$, where $h$ is a user supplied threshold whose selection we will soon discuss, or whenever the denominator of this quantity is 0.  Following change detection, we then update our estimate for the most recent time of change-point occurrence $\tau = t$.  We briefly note this the decision rule \ref{eq:changedetection} bears similarity to the multihypothesis Shiryaev protocol discussed in \cite{tartakovsky1998asymptotic}.  However, while Tartakovsky computed these quantities separately for each possible post-change state, we evaluate a single likelihood ratio whose numerator is averaged across all particles with an alternative post-change state.  In preliminary analysis, we found in the context of our particle filter that such a distinction made little difference, as particles associated with inaccurate post-change states are rapidly filtered out in posterior inference, while being slightly faster to compute.

\

\noindent Implicitly required for the usage of traditional Shiryaev protocols for change detection is the assumption that change-points should be detected one at a time.  This assumption is widespread in the realm of sequential change detection.  While this assumption can in theory be removed, it requires the introduction of considerably more complicated decision rules (see e.g. Tsai et al. \cite{tsai2013optimal} who consider the simultaneous identification of two change-points), and so is largely motivated on the grounds of parsimony.  This assumption further motivates the usage of our indicator variable in our runlength prior Equation \ref{eq:hazard}; If removed, it is plausible that in the context of our inference scheme individual particles could undergo multiple state transitions, i.e., could undergo multiple, short-lived, gradual changes, over short time horizons.  Due to the flexibility of our model and inference scheme such particles may not be filtered, resulting in particles which violate this assumption.  In contrast, by ensuring that particles propagate forward under a null hypothesis in which no new gradual changes have occurred or deviating from this null hypothesis at only a single moment in time this assumption is honored, facilitating the sound computation of our Shiryaev statistic.

\

\noindent Selection of $h$ will dictate a trade off between ADD and PFA.  In the ideal case, where pre and post-change model parameters are known and where changes in process behavior occur abruptly via change-points, we could simply set $h = \frac{1-\alpha}{\alpha}$, where $\alpha$ is a user defined upper bound on an acceptable PFA as is common practice \cite{veeravalli2014quickest}.  As discussed in \cite{lai10}, parameter estimation in tandem with change-detection may necessitate the need for modification to this ideal decision rule.  Though we are unaware of prior works which have considered threshold selection for Shiryaev protocols in the context of particle filter driven inference or in the context of gradual change, motivating such an inquiry as a direction of potential future work, we find in Section \ref{sec:exp} good agreement between our empirical measures of PFA and $\alpha$ for $\alpha$ small when using this choice of threshold.  However, for large $\alpha$, i.e. in regimes where false alarms are more common, we conjecture that our PFA may rise above this theoretical bound, as false alarms leading our change-dynamic model onto an incorrect state sequence may need to be adjusted via the declaration of a second spurious alarm.  Thus, for $\alpha$ large, we propose to use as a lower bound for $h = \frac{1-\frac{\alpha}{2}}{\frac{\alpha}{2}}$, in which we effectively double penalize false alarms to account for their potential correlation.  This potential correlation in false alarms for large $\alpha$ does not seem unique to our model and, although this concern has gone largely undiscussed in related literature, Tartakovsky \cite{tartakovsky1998asymptotic} finds that the agreement between empirical measures of change-detection accuracy break away from theoretical bounds when the probability of post-change state misclassification given change-point declaration tend away from $0$.  As state misclassification corresponds to false alarms in our change-dynamic model, and because these events are more probable for large $\alpha$, this finding seems to support our conjecture.

\subsection{Prediction}
\label{sec:prediction}

\noindent One-step ahead predictions for process values $X_{t+1}$ can be made by sampling from a one-step ahead predictive density.  When generating predictions, we assume no new change-point occurs between times $t$ and $t+1$.  Predictions will thus take the form of a mixture model, whose components correspond to each particle associated with our non change hypothesis at time $t$:

\begin{equation}
    \label{eq:particlepred}
    P(X_{t+1}|\mathbf{x}_{1:t}) \simeq \frac{1}{N}\sum_{i=1}^N I(r_t^i \geq t - \tau) P(X_{t+1}|\mathbf{p}_{1:t}^i,\mathbf{x}_{-p:t}).
\end{equation}

\noindent The mean and variance of this distribution are given:

\begin{subequations}
\label{eq:particlepredmusigma}
\begin{align}
    E[X_{t+1}] = \frac{1}{N}\sum_{i=1}^{N} I(r_t^i \geq t-\tau) \alpha_t^i \tilde{\mathbf{x}}_{t}^i + \mu_{t}^i \\ 
    Var[X_{t+1}] = \sum_{i=1}^{N} I(r_t^i \geq t-\tau) \frac{1}{N} \left( \sigma_i^2 + \alpha_t^i \tilde{\mathbf{x}}_{t}^i + \mu_{t}^i \right) - E^2[X_{t+1}].
\end{align}
\end{subequations}

\noindent  While the mean of this distribution is used to generate our predictive results in experiments, consideration of variance can be attractive in cases requiring uncertainty or risk evaluation.  Analogous expressions for other quantities of interest (e.g. predictions for future process model parameters $\mathbf{\theta}_{t+1}$ or $k$ step ahead predictions) are easily derived, and omitted here in the interest of brevity.

\subsection{Computability}
\label{sec:computability}

The primary computational cost associated with implementation of our change-dynamic model stems from the need to evaluate particle predictive densities $P(X_{t+1}|\mathbf{p}_i^t,\mathbf{x}_{1:t})$ at each time step for use in inference, prediction, and change-detection.  As the cost of these these evaluations is, using rank-one updates \cite{seeger2004low}, $\mathcal{O}(m^2)$, the total computation cost of our method will then be $\mathcal{O}(Nm^2)$ per time step.   Runtime will thus be primarily bottlenecked the by number of particles used for inference.  We do however note that implementation of our model can be parallelized across particles, thereby reducing runtime.

\section{Experiments}
\label{sec:exp}

\subsection{Synthetic Data: Detection of a Single Gradual Transition in Mean}

\noindent To demonstrate how our change-dynamic model can produce faster and more accurate detection of gradual change than change-point baselines, we first consider a synthetic application in which a shift is gradually introduced into the mean of a sequence of Gaussian observations. Data is generated from a Gaussian distribution with time-varying mean:

$$X_{t+1} =  \mu_t +  .05 \epsilon_t,$$

\noindent with $\epsilon_t \sim \mathcal{N}(0,1)$ for $t=1:225$ and with $\mu_0 = 1$.  We note that while this data generating process lacks any autoregressive component, this application is intended to serve as an illustrative example and provide a controlled setting in which we can easily compare our results to well posed change-point baselines, one of which presumes analytical tractability.

\

\noindent During data generation gradual changes in $\mu$ are constructed via a simple, determinstic, linear dynamic:

\begin{equation}
\label{eq:syntrans}
\mu_{t+1} = \mu_t + \nu_{t+1}.
\end{equation}

\noindent We assume a two state system $S \in {0,1}$, with state $S=0$ reserved to indicate periods of model stationarity and with state $S=1$ corresponding to a linear decrease in process mean.  States $S$ are deterministically proposed in data generation, with $S_{t} = 0$ for $t=1:25, t=126:225$, $S_{t} = 1$ for $t=26:125$.  Conditional on state, we also deterministically degenerate  $\nu_t | S_t = 0 \sim \delta(\cdot;0), \nu_t | S_t = 0 \sim \delta(\cdot;-.002)$.

\

\noindent In selection of a process model for this task, we assume the appropriate AR(0) specification of process dynamics.  We further assume knowledge of $\mu_0$, of the process standard deviation $.05$, and of $s_0 = 0$.  As there are only two states in this model, our probability transition matrix $P_S$ must be the matrix with ones on the off diagonal.  We take $p(0) = \frac{1}{25}, p(1)=\frac{1}{100}$.  However, in an effort to incorporate model uncertainty, we replace the deterministic change-dynamic of Equation \eqref{eq:syntrans} with a noisy variant:
$\mu_{t+1} = \mu_t + \nu_{t+1} + \gamma_{t+1} \omega_{t}$, with $\omega_t \sim \mathcal{N}(0,1)$.  We take $\mathbf{\eta}_0^{min} = [0;.0001],\mathbf{\eta}_0^{max} = [0;.001],\mathbf{\eta}_1^{min} = [-.0022;.0001],\mathbf{\eta}_1^{max} = [-.018;.001]$.  These bounds over $\nu$ were chosen to correspond to an assumption that the rate of gradual change in $\mu$ is specified to within 10 percent of its true value.  During preliminary testing, we found that wider bounds produced almost identical results, but with higher detection delay.  Bounds over $\gamma$ were chosen manually.  However, we found our results to be highly robust across different bounds, provided the upper bound on $\gamma < .01$.  In contrast, for $\gamma \geq .01$, our model struggled to differentiate between states as either state could fit changes in $\mu$ well.  This is unsurprising, as in this regime the random walk strength of our change-dynamic effectively dominates the ground truth rate of change in $\nu$.

\

\noindent We compare our change-dynamic model to two baselines: First, a self implemented two-state Bayesian change-point model, in which inference is performed identically to our change-dynamic model, but with $\mu$ assumed to abruptly jump at change-points: $\mathbf{\mu}_t | S_t = 0 \sim \delta(\cdot;0), \mathbf{\mu}_t | S_T = 1 \sim \delta(\cdot;.8)$.  We note this baseline has exact knowledge of pre and post transition model parameters, and should be expected to perform relatively well.  Second, an implementation of the Bayesian Online Changepoint Detection (BOCD) algorithm of Adams, Mackay \cite{adams2007bayesian}.  Our change-dynamic and change-point models are fit using $N=2000$ particles.  As motivated in Section \ref{sec:changedetection}, we fit these methods using thresholds of $h=19,99$, chosen to optimistically conform with theoretically ideal upper bounds on PFA of $5$ and $1$ percent respectively.  In application of BOCD we took a Gaussian prior over $\mu \sim \mathcal{N}(1,.03)$ and a gamma prior over the precision $\tau \sim Gamma(.001,.001)$.  We found the results of this method to be extremely robust with respect to adjustment in the hyperparameters of these distributions.  For BOCD, we take a geometric prior over $R \sim Geom(\frac{2}{125})$ with rate equal to the average rate of ground truth change in our system, for which results were also quite robust.  For this competitor we declare a change has occurred whenever the posterior over change-point occurrence exceeds a given threshold and report results for thresholds correspondin to an uppder bound on PFA of $5$ and $1$ percent.  We note that the PFA and predictive accuracy of BOCD, our change-point baseline, and our change-dynamic model were quite insensitive to choice of threshold $h$ for $h \geq 19$, but with higher thresholds producing heightened ADD for all models.

\

\noindent We report model fit as measured by comparing the RMSE between our one-step ahead mean posterior predictions for $\mu$ and true values (RMSFE).  For our change-dynamic model, we also report RMSE between the final model posterior over $\nu$ and its true values to assess whether or not our model can reliably infer the rate of this gradual change.  To compare the reliability of these models for change-detection, we report PFA as the ratio of alarms which were raised before change-point occurrence or with inaccurate post-change state to total alarm count.  ADD, measured as the time between change-point occurrence and the earliest alarm raised after said time with appropriate post-change state, is reported separately for each of the two ground truth change-points for reasons that will soon be made clear.  We also report the number of alarms $A$ raised each method.

\

\begin{figure}[H]
\centering
\includegraphics[scale=0.5]{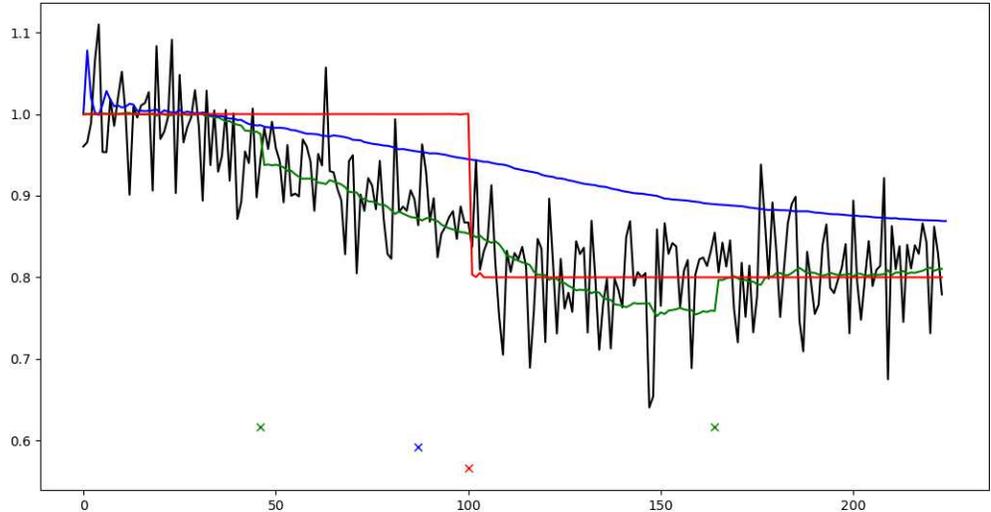}
 \caption{One-step Ahead Mean Predictions for Change-dynamic (green), Change-point (red), and BOCD (blue) of $\mu$ vs One Realization of True Data (black). Green, red and blue marks indicate times of change detection for each model with $h=99$.  Ground truth changes occur at $t=25,125$.}
\label{fig:syn1preds}
\end{figure}

\noindent In Table \ref{tab:synthetic1} we report these summary statistics, while displaying in Figure \ref{fig:syn1preds} illustrative predictive results. We immediately see our change-dynamic model outperforms competitors in prediction of $\mu$, due to its ability to properly model gradual changes in process behavior.  We also note relatively poor predictive performance associated with BOCD.  This finding is likely explained by the extent to which this model's posterior over $\mu_t$, for $t$ after gradual change termination, is effectively tainted by pre-change data or data sampled towards the beginning of gradual change.  This causes any hypothetical post-change model to be biased towards pre-change process behavior, as previously discussed.  We note again these results are robust to hyperparameter selection.

\begin{table}[H]
\begin{tabular}{|c|c|c|c|c|c|c|}\hline
  & RMSFE($\mu$)  & RMSE($\nu$) & ADD1 & ADD2 & PFA & PMA \\ \hline   
Change-dynamic $h=19$ & .05 & .0008 & 22 & 30 & .03 & 0 \\ \hline 
Change-dynamic $h=99$ & .05 & .0008 & 23 & 36 & 0 & 0 \\ \hline 
Change-point  $h=19$ & .07 & NA & 60 & FTD & 0 & .5 \\ \hline     
Change-point  $h=99$ & .07 & NA & 65 & FTD & 0 & .5 \\ \hline 
BOCD $h=19$ & .09 & NA & 53 & FTD & 0 & .5 \\ \hline
BOCD $h=99$ & .09 & NA & 61 & FTD & 0 & .5 \\ \hline
\end{tabular}
\caption{Performance Metrics Indexed by Methodology.  Reported metrics are averaged across 30 independent trials.  ADD1, ADD2 correspond to the average detection delays for gradual change onset and termination respectively.  FTD indicates a failure to detect a given change.  We note this failure to detect change occurs across all realizations of our data.  We have omitted entries for RMSE($\nu$) for change-point competitors, as this quantity is not considered by these models.}
\label{tab:synthetic1}
\end{table}

\

\noindent  Of greater interest is how these models differ in their treatment of gradual change.  Observing Figure \ref{fig:syn1preds}, we see that within our change-dynamic model two distinct change-points are identified corresponding to the onset of a gradual transition and its corresponding termination.  In contrast, both our change-point baseline and BOCD identify a single change-point whose timing corresponds roughly to the midpoint of our gradual transition.  Moreover, the detection delay associated with this change-point is considerably higher than e the detection delay associated with identification of gradual change onset by our change-dynamic model.  We note this observation is robust across realizations of our process.  Thus, in this example, traditional change-point models fail to properly identify the number and timing of ground truth changes in model behavior, instead detecting the onset of gradual change with an unacceptably high detection delay while preemptively declaring the end of gradual change.  While this clearly impacts the accuracy of process value predictions during this transitory period, the failure of these methods to appropriately signal an end to a gradual change could be quite problematic in online settings where a control may not be turned on or shut off at the appropriate time.  This preemptive declaration of gradual change termination is our motivation in reporting detection delay separately for each underlying change-point, as the change-point baselines effectively declare the occurrence of gradual change termination prematurely and hence have no well posed detection delay.

\

\begin{figure}[H]
\centering
\begin{subfigure}{.48\textwidth}
\centering
\includegraphics[width=\textwidth,clip=false]{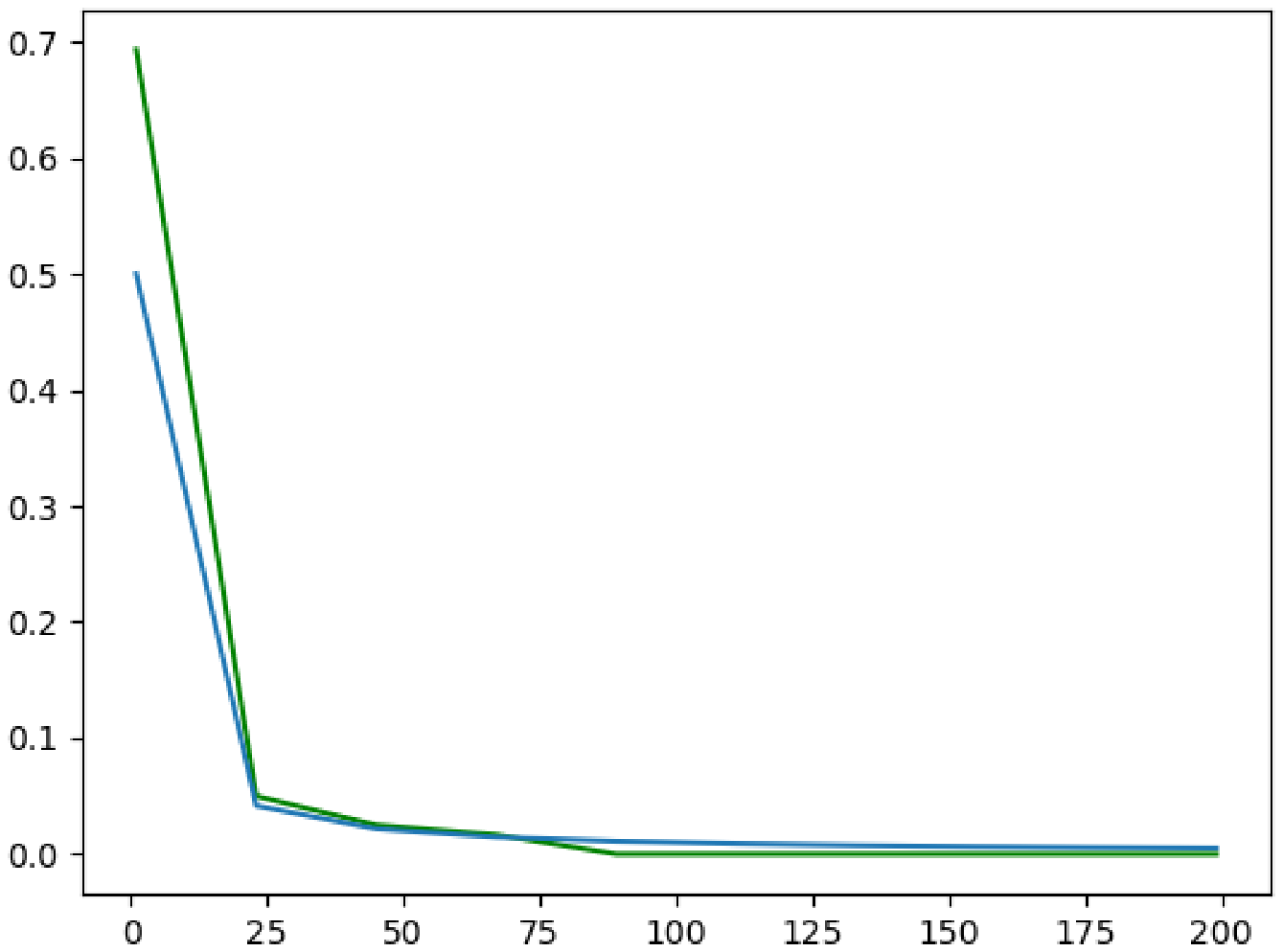}
\end{subfigure}
\begin{subfigure}{.48\textwidth}
\centering
\includegraphics[width=\textwidth,clip=false]{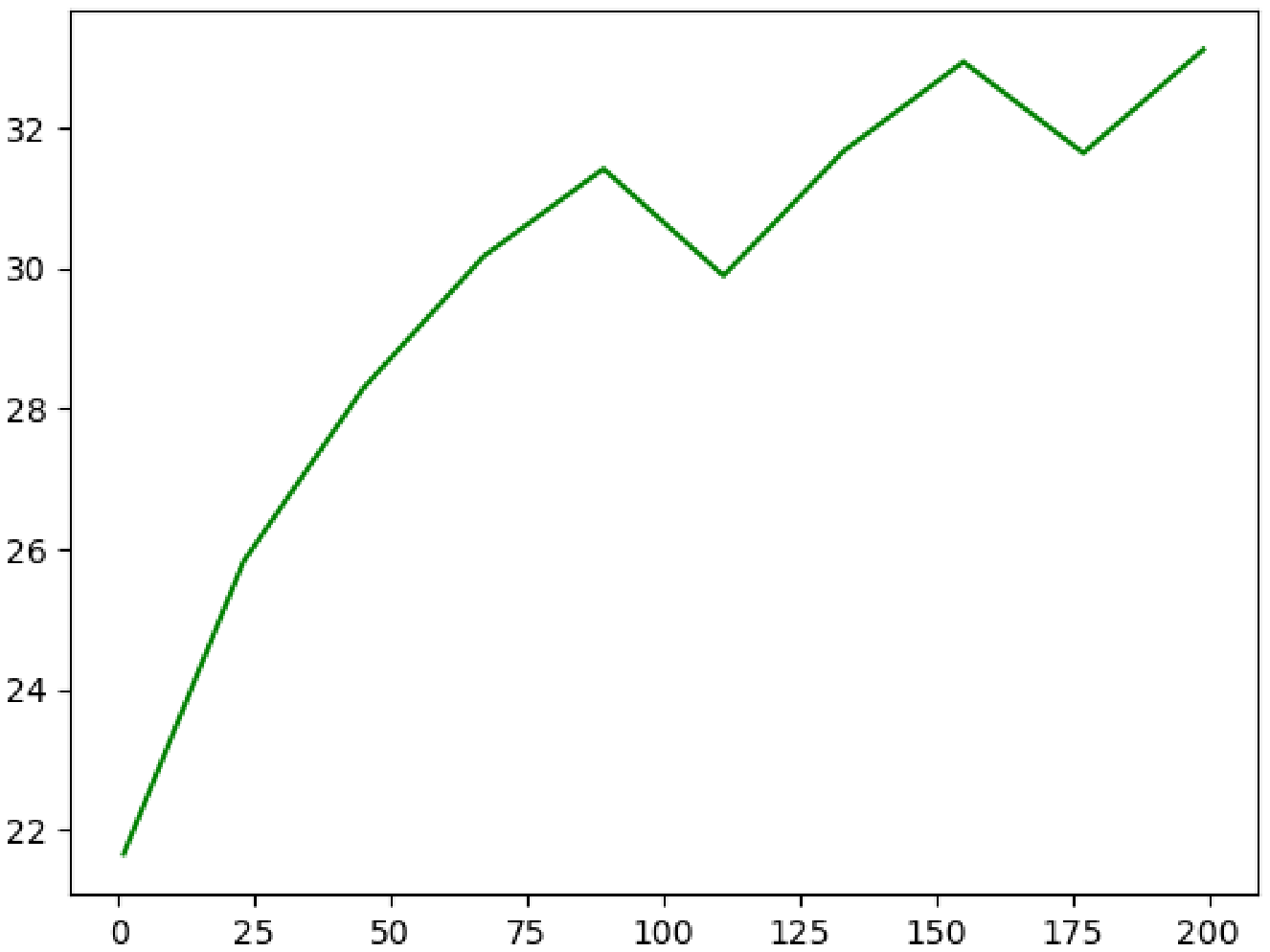}
\end{subfigure}
\caption{(Left): PFA as function of $h$ for change-dynamic model (green) against theoretical upper bound (blue).  (Right): ADD as function of $h$ for change-dynamic model.  Statistics are computed jointly across both change-points and averaged over 30 independent trials for each threshold.}
\label{fig:faraddvsh}
\end{figure}

\noindent  Despite our lack of theoretical guarantees on PFA due to our online estimation of model parameters and approximation of a traditional Shiryaev protocol via a particle filter, we note that with $h=19$ our change-dynamic model reports a $PFA \simeq 3$ percent, corresponding to one false alarm raised in one realization of our process.  While below a theoretical bound of $5$ percent, we note our change-point competitors reported $0$ false alarms in this setting.  In contrast, using a considerably higher threshold $h=99$ we attain a PFA of $0$ percent, below a theoretical bound of $1$ percent.  In section \ref{sec:changedetection}, we conjectured that false alarms may occur more frequently than expected for small $h$ (large $\alpha$) due to the need for a second false alarm to correct failures in our model.  To test this conjecture and allow us to empirically examine the rate of decay in PFA as function of threshold, we present in Figure \ref{fig:faraddvsh} both PFA and ADD for small $h$ linearly spaced between $1$ and $199$, where false alarms are more likely to be present.  For $h$ small, we do find the PFA of our change-dynamic model to be higher than theoretically expected, and in particular find this discrepancy to be quite significant when $h=1$.  We note that by instead applying our modified threshold of $h = \frac{1-\frac{\alpha}{2}}{\frac{\alpha}{2}}$ such discrepancy would be avoided, suggesting this heuristic may be useful when gradual change detection is performed in scenarios where low ADD is preferable to higher PFA.

\

\begin{figure}[H]
\centering
\includegraphics[scale=0.5]{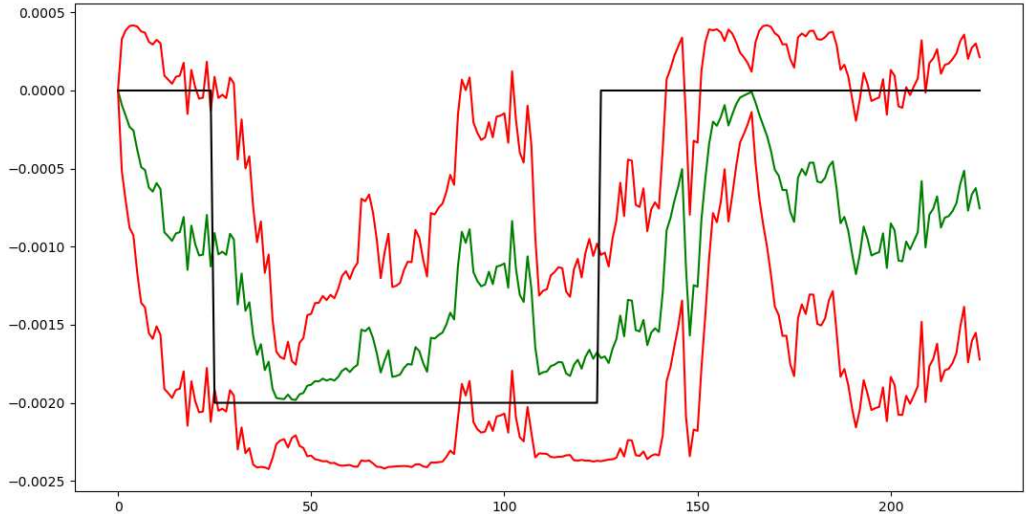}
\caption{Mean (green) +/- one standard deviation (red) of posterior over $\nu$ for change-dynamic model with $h=99$ against true values (black).  }
\label{fig:syn1phi}
\end{figure}

\noindent In Figure \ref{fig:syn1phi}, we present our final posterior (at time $t=225)$ over the rate of change $\nu$ in process mean $\mu$ to more fully assess the efficacy of our model for the inference of the rate at which a system gradually changes over time.  Noting that ground truth values for $\nu$ remain within one standard deviation of our model predictions, but typically deviate away from our exact mean predictions within this range, we can see that our model exhibits moderate but not overwhelming success in inferring the rate of gradual change in process behavior.   Given the flexibility of our model, this is not too surprising, as small deviations in the rate of change $\nu$ in process mean can be alternatively described via direct modifications to the posterior over $\mu$ or $X$.  Of particular interest is a marked reduction in both posterior uncertainty and marked increased in posterior accuracy (i.e. RMSE between posterior mean and true value) at times of change-detection, where our posterior effectively collapses onto the ground truth value of $\nu$.  This high posterior precision at times of change-detection indicates that our model may produce its most reliable estimates for the rate of a gradual change at the moment it is detected, and thus these moments could be used in practice to most accurately infer the rate of a given gradual change.

\subsection{Seizure Detection}

\noindent  Seizures are transient periods of aberrant neural activity which cause temporary loss of motor control and altered consciousness.  Prior to seizure onset, epileptics are advised to take care to place themselves in a safe environment so as to avoid injury prior to seizure onset.  When injury does occur, it may be challenging for the incapacitated individual to call for help.  As these events are challenging to predict in advance, the rapid identification of seizure onset via electronic monitoring devices can thus be of great benefit to epileptics, by alerting third parties who can ensure the epileptic safety.  Moreover, as seizures which do not naturally remiss within a short time frame (status epilepticus) require more serious medical intervention, automated online inference of the rate at which seizure onset and remission is or is not occurring could be useful in automating the dispatch of medical personnel to aid in potentially life threatening situations.

\

\begin{figure}
    \centering
    \includegraphics[scale=0.5]{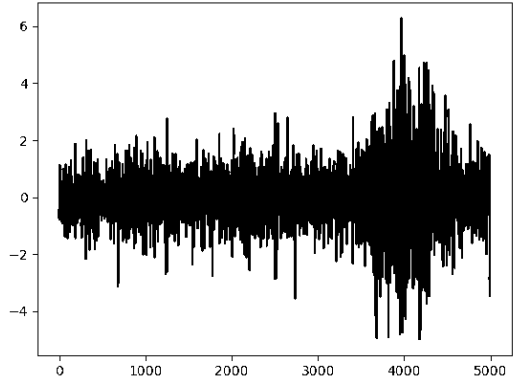}
    \caption{EEG Readings Throughout a Focalized Temporal Seizure.}
    \label{fig:raweegdata}
\end{figure}

\noindent Both seizure onset and termination can occur gradually, with seizures typically comprised of distinct pre-ictal, ictal, and post-ictal states. This gradual onset can pose challenges for detection protocols which compare current data to a recent memory of readings \cite{gotman1990automatic}.  By explicitly considering the gradual onset and termination of epileptic events, we aim to potentially lower the delay of seizure detection over change-point baselines, while providing indications of times where a seizure seems to be remissing or growing worse.  We thus consider a dataset\cite{ombao2001automatic,ombao2005slex} consisting of EEG readings drawn just prior to and throughout the duration of a focal seizure originating in the temporal lobe.  While our original data consists of $50000$ continuous EEG readings, sampled at 100hz from a sensor placed near the temporal lobe of the patient, we chose to downsample these readings to 10hz for computational reasons, giving a time series of length $T=5000$ to which a low pass filter of 5hz was subsequently applied so as to ensure accurate signal representation.  Data was then normalized to have 0 mean and unit standard deviation for convenience.  In Figure \ref{fig:raweegdata} we display our processed data.  We note the first $~3550$ and last $~600$ datums seem to correspond to pre and post ictal (seizure) behavior respectively, while the remaining observations coinciding with predominantly ictal activity.  Given that this ictal activity occurs only in the later portions of our data, we use only the last 2000 observations as testing data to evaluate our model, while using the first 3000 observations for training purposes.

\

\noindent  While a more nuanced model of seizure dynamics might incorporate distinct pre-ictal, ictal, and post-ictal states, we take a simpler approach motivated by our relative dirth of training data.  We adopt a simple three state model, with state $S=0$ again corresponding to periods of process stationarity, state $S=1$ corresponding to a gradual transition into ictal activity, and state $S=2$ corresponding to a gradual transition out of ictal activity.  As linear models are known to describe EEG readings sampled during epileptic events poorly relative to EEG readings sampled during non-ictal periods \cite{ombao2005slex} we consider a simple AR(2) process model:

\begin{equation}
    x_t =  \alpha_1 x_{t-1} + \alpha_2 x_{t-2} + \sigma_t \epsilon_t 
\label{eq:eegpm}
\end{equation}

\noindent where $\epsilon_t \sim \mathcal{N}(0,1) \ \forall t$, and aim to identify seizure onset through identification of gradual changes in the magnitude of residuals about this linear model.  We estimate $\theta_1 = .62, \theta_2 = -.18$ from our training data, with $\log(\sigma_0) = -1$ as our corresponding log RMSE on said training data.  We note this model performs surprisingly well in modeling non-ictal EEG dynamics, with an $r^2 \simeq .75$ on our training data alongside an $r^2 \simeq .49$ when evaluated on our ``post-ictal" data (final 600 observations of testing data).

\

\noindent In designing our change dynamic model for $\sigma$: 
$\log{(\sigma_{t+1}}) = \log{(\sigma_{t})} + \nu_{t+1} + \gamma_{t+1} \omega_t$, where $\omega_t \sim \mathcal{N}(0,1) \ \forall t$, selection of our uniform bounds over $\nu$ for each state was performed as follows:  First, we computed the mean standard error about our process model during peak ictal activity (testing datums 900 through 1200) where we found $\log(\sigma) \simeq 1$.  We then combined this observation with prior knowledge that ictal events born of focalized, temporal lobe, epilepsy typically last between 30 to 120 seconds, which in our model would correspond to seizures lasting roughly $300$ to $1200$ timesteps.  By making a naive assumption that transition time from non ictal to ictal activity is equal in duration to the time to needed to revert back to non ictality from ictality, we can then bound the rate of change in $\log(\sigma)$ by $[\frac{2}{600},\frac{2}{150}]$. We thus take $\eta_1^{min} =[.0033;.0003], \eta_1^{max} = [.013;.0003], \eta_2^{min} = [-.013;.0003], \eta_2^{max} = [-.0033;.0003]$.  We note that the use of testing data to define this prior is non-ideal.  However, given that we have only one realization of our process, such a procedure seems defensible.  We further note this information gained from testing data is used only to inform our prior over seizure intensity and not duration.  To allow for slight heteroskedastic adjustment during periods of model stationarity, we take $\eta_0^{min} = [-.0033;.0003], \eta_0^{max} = [.0033;.0003]$.  Our prior bounds over $\gamma$, here assumed to be identical across states, were chosen manually to be one order of magnitude smaller than those over $\nu$.  Again we found our model performance to be insensitive variations in these prior bounds for $\gamma$ small, e.g. $\gamma \leq .0001$.  Again we found for larger $\gamma \geq .001$ that states became functionally indistinguishable as the strength of this random walk allowed for the accurate reconstruction of heteroskedasticity without the need for changes in state.  We take $p(S) = [\frac{1}{10000},\frac{1}{375},\frac{1}{375}]$, with $\frac{1}{375}$ taken to coincide with the mean of our uniform prior over half seizure runlengths and with $\frac{1}{10000}$ simply chosen to encode a prior belief that transitions out of model stationarity should be relatively rare, though still likely enough that a sufficient number of particles will enter this state during prior proposals.

\

\noindent In comparison to our self implemented change-point baseline, we again employ a two state model with $\log(\sigma_t) | S_t = 0 = -1$, $\log(\sigma_t) | S_t = 1 = 1$.  In comparison to BOCD, we treat model residuals as Gaussian with mean 0 and unknown precision $\tau \sim Gamma(.001,.001)$.  We set $\lambda = \frac{1}{10000}$
as the mean of BOCD's runlength prior to ensure similar seizure onset detection times as our change-dynamic model.  We note that BOCD model performance was relatively insensitive to hyperparameter selection for our prior over $\tau$. Given our assumption that transitions out of stationarity model behavior occur infrequently, we fit our change-dynamic and change-point models with $N=5000$ particles to ensure sufficient posterior coverage.  Motivated by our empirical results in Section 4.1, we use a shared threshold of $h=99$ for all models.

\

\begin{figure}[H]
\centering
\includegraphics[scale=0.5]{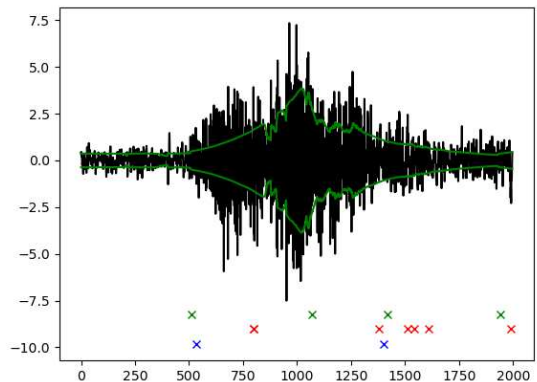}
\caption{Testing Residuals About Baseline AR(2) Model (Black) VS +- Change-dynamic Posterior Mean Over $\sigma$ (Green).  Green, red, and blue x marks indicate alarms raised by change-dynamic, change-point and BOCD models respectively. }
\label{fig:eegpreds}
\end{figure}

\noindent Given our treatment of this problem via heteroskedasticity about a linear model unintended to accurately predict ictal EEG dynamics, we emphasize that we do not consider predictive accuracy to be a particularly relevant metric for model performance in this case.  Moreover, due to the unavailability of ground truth data marking change-point locations, we are unable to quantitatively assess the performance of our change-dynamic model or baselines in detecting change in this setting.  Instead, we compare qualitatively the results of these methods for change-detection.  In Figure \ref{fig:eegpreds}, we display marked times of change detection for each model alongside our final posterior over $\sigma$ against residuals.  Of immediate interest is that while BOCD and our change-dynamic model declare initial change (i.e. seizure onset) at relatively similar times, our change-dynamic model identifies this change roughly two seconds earlier than BOCD.  In contrast, our change-point baseline seems to exhibit high ADD, declaring seizure onset roughly halfway to seizure peak.  Although we concede that in practice the two seconds gained through consideration of seizure gradual onset may or may not make a large difference in the management of epilepsy, this slightly earlier alarm could be beneficial in some scenarios, giving epileptics slightly more time to anticipate seizure onset or facilitating slightly faster administration of aid.  Moreover, this finding continues to reinforce our core narrative that our change-dynamic model is capable of detecting gradual change faster than change-point baselines.  

\

\

\noindent Beyond this first change-point, we see that BOCD raises one additional alarm coinciding with the end of ictal activity.  This alarm is raised roughly two seconds earlier than a similar alarm raised by our change-dynamic model.  Both our change-dynamic model and change-point baseline declare an alarm coinciding with a large EEG spike at the end of observation.  We conjecture this alarm may be associated with muscle or eye movements coinciding with the end of observation, and note by visual inspection this alarm seems quite reasonable to raise.  Our change-dynamic model then raises one additional alarm, which is not identified by BOCD or our change-point baseline, which seems to coincide with peak ictal activity, i.e., the transition point between seizure onset and termination.  This accurate identification of peak ictality seems desireable.  As earlier discussed,  seizures which do not remiss within an appropriate timeframe typically require more serious medical intervention by third parties.  As our change-dynamic model, in contrast to change-point baselines, seems able to identify the transition from seizure onset to seizure termination, failure by our model to raise this alarm within a reasonable timeframe could theoretically serve as an automated method for detecting the onset of status epilepticus or incidence of seizure clusters.

\section{Conclusion and Future Work}

Gradual change is a widespread phenomenon in the natural world.  Despite this observation, the overwhelming majority of prior work in change-detection has treated change occurrence as instantaneous and complete.  In this work, we introduced a novel change-dynamic model for the online detection of gradual change in an effort to produce faster and more accurate detection of such events.  On simulated data, we found empirically that this model can produce faster and more accurate detection of gradual change than comparable change-point baselines, while allowing for modestly accurate inference of the rate of these changes. We then showed how this model could potentially be employed to aid in the management of epilepsy.  Although we considered here a relatively simple description of gradual process change, in which gradual changes in the parameters defining a sequence of univariate Gaussian observations underwent exclusively linear transitions over time, our usage of a particle filter for approximate inference allows for the straightforward extension of our work to more complex process models and change-dynamics, which could serve as a basis for future development.  For modestly large thresholds we found the probability of false alarm in our change-dynamic model remained below its theoretical bound, in contrast to our findings for small thresholds.  Thus, on the theoretical side, a more rigorous analysis on the performance of CUSUM or Shiryaev-protocols used in the context of particle filter driven inference and/or gradual change-detection could be of interest.

\end{document}